\documentclass[11pt]{article}
\usepackage{naaclhlt2013}
\usepackage{times}
\usepackage{latexsym}
\usepackage{amsmath}
\usepackage{multirow}
\usepackage{arabtex}
\usepackage{url}
\usepackage{pgfplots}
\usepackage{pbox}

\title{Arabizi Detection and Conversion to Arabic}

\author{Kareem Darwish \\
  Qatar Computing Research Institute \\
  Qatar Foundation, Qatar \\
  {\tt kdarwish@qf.org.qa} \\
}

\date{}

\begin{document}
\setarab
\novocalize
\maketitle

\begin{abstract}
Arabizi is Arabic text that is written using Latin characters.  Arabizi is used to present both Modern Standard Arabic (MSA) or Arabic dialects. It is commonly used in informal settings such as social networking sites and is often with mixed with English.  In this paper we address the problems of: identifying Arabizi in text and converting it to Arabic characters. We used word and sequence-level features to identify Arabizi that is mixed with English.  We achieved an identification accuracy of 98.5\%.  As for conversion, we used transliteration mining with language modeling to generate equivalent Arabic text. We achieved 88.7\% conversion accuracy, with roughly a third of errors being spelling and morphological variants of the forms in ground truth.
\end{abstract}

\section{Introduction}
\label{sect:intro}
Arabic is often written using Latin characters in transliterated form, which is often referred to as Arabizi, Arabish, Franco-Arab, and other names.  Arabizi uses numerals to represent Arabic letters for which there is no phonetic equivalent in English or to account for the fact that Arabic has more letters than English.  For example, ``2" and ``3" represent the letters <'a> (that sounds like ``a" as in apple) and <`> (that is a guttural ``aa") respectively.  Arabizi is particularly popular in Arabic social media.  Arabizi has grown out of a need to write Arabic on systems that do not support Arabic script natively.  For example, Internet Explorer 5.0, which was released in March 1999, was the first version of the browser to support Arabic display natively\footnote{\url{http://en.wikipedia.org/wiki/Internet_Explorer}}.  Windows Mobile and Android did not support Arabic except through third party support until versions 6.5x and 3.x respectively.  Despite the increasing support of Arabic in many platforms, Arabizi continues to be popular due to the familiarity of users with it and the higher proficiency of users to use an English keyboard compared to an Arabic keyboard.  Arabizi is used to present both MSA as well as different Arabic dialects, which lack spelling conventions and differ morphologically and phonetically from MSA.  Additionally, due to the fact that many of the Arabic speakers are bilingual (with their second language being either English or French), another commonly observed phenomenon is the presence of English (or French) and Arabizi mixed together within sentences, where users code switch between both languages.  In this paper we focus on performing two tasks, namely: detecting Arabizi even when juxtaposed with English; and converting Arabizi to Arabic script regardless of being MSA or dialectal.  Detecting and converting Arabizi to Arabic script would help:  ease the reading of the text, where Arabizi is difficult to read; allow for the processing of Arabizi (post conversion) using existing NLP tools; and normalize Arabic and Arabizi into a unified form for text processing and search.   Detecting and converting Arabizi are complicated by the following challenges:
\begin{enumerate}
\item
Due to the lack of spelling conventions for Arabizi and Arabic dialectal text, which Arabizi often encodes, building a comprehensive dictionary of Arabizi words is prohibitive.  Consider the following examples:
\begin{enumerate}
	\item
		The MSA word <t.hryr> (liberty) has the following popular Arabizi spellings: ta7rir, t7rir, tahrir, ta7reer, tahreer, etc.
	\item
		The dialectal equivalents to the MSA <lA yl`b> (he does not play) could be <mAbyl`b^s>, <mAbl`b^s>, <myl`b^s>, <mAyl`b^s> … etc.  The resultant Arabizi could be: mayel3absh, mabyelaabsh, mabyel3absh, … etc.
\end{enumerate}
\item
Some Arabizi and English words share a common spelling, making solely relying on an English dictionary insufficient to identify English words.  Consider the following examples (ambiguous words are bolded):
\begin{enumerate}
	\item
		Ana 3awez aroo7 \textbf{men} America leh Canada (I want to go from America to Canada). The word ``men" meaning ``from" is also an English word.
	\item
		I called \textbf{Mohamed} last night.  ``Mohamed" in this context is an English word, though it is a transliterated Arabic name.
\end{enumerate}
\item
Within social media, users often use creative spellings of English words to shorten text, emphasize, or express emotion.  This can complicate the differentiation of English and Arabizi.  Consider the following examples:
\begin{enumerate}
	\item
		I want 2 go with u tmrw, cuz my car is broken.
	\item
		Woooooow. Ur car is cooooooool.
	\end{enumerate}
\end{enumerate}

Due to these factors, classifying a word as Arabizi or English has to be done in-context.  Thus, we employed sequence labeling using Conditional Random Fields (CRF) to detect Arabizi in context.  The CRF was trained using word-level and sequence-level features.  For converting Arabizi to Arabic script, we used transliteration mining in combination with a large Arabic language model that covers both MSA and other Arabic dialects to properly choose the best transliterations in context.

The contributions of this paper are:
\begin{itemize}
\item We employed sequence labeling that is trained using word-level and sequence-level features to identify in-sentence code-switching between two languages that share a common alphabet.
\item We used transliteration mining and language modeling to convert form Arabizi to Arabic script.
\item We plan to publicly release all our training and test data.
\end{itemize}

The remainder of the paper is organized as follows:  Section 2 provides related work; Section 3 presents our Arabizi detection and reports on the detection accuracy; Section 4 describes our Arabizi to Arabic conversion approach and reports the accuracy of conversion; and Section 5 concludes the paper.

\section{Related Work}
\label{sect:related}
There are two aspects to this work: the first is language identification, and the second is transliteration.
There is much work on language identification including open source utilities, such as the Language Detection Library for Java\footnote{\url{http://code.google.com/p/language-detection/}}.  Murthy and Kumar~\shortcite{MurthyKumar2006} surveyed many techniques for language identification.  Some of the more successful techniques use character n-gram models~\cite{Beesley1988,Dunning1994} in combination with a machine learning technique such as hidden Markov models (HMM) or Bayesian classification~\cite{Xafopoulosetal2004,Dunning1994}.  Murthy and Kumar~\shortcite{MurthyKumar2006} used logistic regression-like classification that employed so-called ``aksharas" which are sub-word character sequences as features for identifying different Indian languages.  Ehara and Tanaka-Ishii~\shortcite{EharaTanaka2008} developed an online language detection system that detects code switching during typing, suggests the language to switch to to the user, and interactively invokes the appropriate text entry method.  They used HMM based language identification in conjunction with an n-gram character language model. They reported up to 97\% accuracy when detecting between two languages on a synthetic test set.  In our work, we performed offline word-level language identification using CRF sequence labeling, which conceptually combines logistic regression-like discriminative classification with an HMM-like generative model~\cite{Laffertyetal2001}.  We opted to use a CRF sequence labeling because it allowed us to use both state and sequence features, which in our case corresponded to word- and sequence-level features respectively.  One of the downsides of using a CRF sequence labeler is that most implementations, including CRF++ which was used in this work, only use nominal features.  This required us to quantize all real-valued features.  

Converting between from Arabizi to Arabic is akin to transliteration or Transliteration Mining (TM).  In transliteration, a sequence in a source alphabet or writing system is used to generate a phonetically similar sequence in a target alphabet or writing system.  In TM, a sequence in a source alphabet or writing system is used to find the most similar sequence in a lexicon that is written in the target alphabet or writing system. Both problems are fairly well studied with multiple evaluation campaigns, particularly at the different editions of the Named Entities Workshop (NEWS)~\cite{Zhangetal2011,Zhangetal2012}.  In our work we relied on TM from a large corpus of Arabic microblogs.  TM typically involves using transliteration pairs in two different writing systems or alphabets to learning character (or character-sequence) level mappings between them.  The learning can be done using the EM algorithm~\cite{Kuoetal2006} or HMM alignment~\cite{Udupaetal2009}. Once these mappings are learned, a common approach involves using a generative model that attempts to generate all possible transliterations of a source word, given the character mappings between two languages, and restricting the output to words in the target language~\cite{ElKahkyetal2011,NoemanMadkour2010}.  Other approaches include the use of locality sensitive hashing~\cite{UdupaKumar2010} and classification~\cite{Jiampojamarnetal2010}.  Another dramatically different approaches involves the unsupervised learning of transliteration mappings from a large parallel corpus instead of transliteration pairs~\cite{Sajjadetal2012}.  In our work, we used the baseline system of El-Kahky et al.~\shortcite{ElKahkyetal2011}.  There are three commercial Input Method Editors (IMEs) that convert from Arabizi to Arabic, namely: Yamli\footnote{\url{http://www.yamli.com/editor/ar/}}, Microsoft Maren\footnote{\url{http://www.getmaren.com}}, and Google t3reeb\footnote{\url{http://www.google.com/ta3reeb}}.  Since they are IMEs, they only work in an interactive mode and don't allow for batch processing. Thus they are difficult to compare against. Also, from interactively using Arabic IMEs, it seems that they only use unigram language modeling.

\section{Identifying Arabizi}
As mentioned earlier, classifying words as English or Arabizi requires the use of word-level and sequence-level features.  We opted to use CRF sequence labeling to identify Arabizi words.  We used the CRF++ implementation with default parameters~\cite{ShaPereira2003}.  We constructed training and test sets for word-level language classification from tweets that contain English, Arabizi, or a mixture of English and Arabizi.  We collected the tweets in the following manner:
\begin{enumerate}
\item
We issued commonly used Arabizi words as queries against Twitter multiple times.  These words were ``e7na" (we), ``3shan" (because), and ``la2a" (no).  We issued these queries every 30 seconds for roughly 1 hour.  We put large time gaps between queries to insure that the results were refreshed.
\item
We extracted the user IDs of all the authors of the tweets that we found, and used the IDs as queries to Twitter to get the remaining tweets that they have authored.  Our intuition was that tweeps who authored once in Arabizi would likely have more Arabizi tweets.  Doing so helped us find Arabizi tweets that don't necessarily have the aforementioned common words and helped us find significantly more Arabizi text. In all we identified 265 tweeps who authored 16,507 tweets in the last 7 days, containing 132,236 words.  Of the words in the tweets, some of them were English, but most of them were Arabizi.
\end{enumerate}

\begin{table}[h]
\begin{center}
{\scriptsize
\begin{tabular}{|c|p{6cm}|}
\hline
Label & Explanation\\\hline
a & Arabizi\\\hline
e & English\\\hline
o & Other including URL's, user mentions, hashtags, laughs (lol, ☺, :P, xd), and none words \\\hline
 \end{tabular}
}
\end{center}
\caption{\label{labels}Used labels for words}
\end{table}

We filtered tweets where most of the words contained Arabic letters.  As in Table~\ref{labels}, all the tokens in the set were manually labeled as English (``e"), Arabizi (``a"), or other (``o"). For training, we used 522 tweets, containing 5,207 tokens.  The breakdown of tokens is: 3,307 English tokens; 1,203 Arabizi tokens; and 697 other tokens.  For testing, we used 101 tweets containing 3,491 tokens.  The breakdown of the tokens is: 797 English tokens; 1,926 Arabizi tokens; and 768 other tokens.  Though there is some mismatch in the distribution of English and Arabizi tokens between training and test sets, this mismatch happened naturally and is unlikely to affect overall accuracy numbers.
For language models, we trained two character level language models: the first using 9.4 million English words; and the second using  1,000 Arabizi words (excluding words in the test set).  We used the BerkeleyLM language modeling toolkit.

We trained the CRF++ implementation of CRF sequence labeler using the features in Table~\ref{features} along with the previous word and next word.  The Table describes each feature and shows the features values for the word ``Yesss".

\begin{table*}[!ht]
\begin{center}
{\scriptsize
\begin{tabular}{|p{1.7cm}|p{13cm}|p{0.5cm}|}
\hline
Feature	&	Explanation	&	Ex.	\\\hline
Word	&	This would help label words that appear in the training examples.  This feature is particularly useful for frequent words.	&	yesss	\\\hline
Short	&	This would remove repeated characters in a word.  Colloquial text such as tweets and Facebook statuses contain word elongations.	&	yes	\\\hline
IsLaugh	&	This indicates if a word looks like a laugh or emoticon.  For example lol, J, :D, :P, xD, (ha)+, etc.  Smiles and laughs should get an ``o" label.	&	0	\\\hline
IsURL	&	This indicates if a token resembles as URL of the form: \verb/http://[a-zA-z0-9\.\/]+/.  URLs should get an ``o" label.	&	0	\\\hline
IsNo	&	This indicates if a token is composed of numbers only.  Numbers should get an ``o" label	&	0	\\\hline
Is!Latin	&	This indicates if a word is composed of non-Latin letters.  If a word is composed on non-Latin characters, it is not ``e".	&	0	\\\hline
WordLength	&	This feature is simply the token length.  Transliterated words are typically longer than native words	&	8	\\\hline
IsHashtag	&	This indicates if it is a hashtag. Hashtags would get an ``e" label.	&	0	\\\hline
IsNameMention	&	This indicates if it is a name mention. Name mentions, which start with ``@" sign, should get an ``o" label.	&	0	\\\hline
IsEmail	&	This indicates if it is an email. Emails, which match \verb/[\S\.\-_]+@[\S\.\-_]+/ should get an ``o" label.	&	0	\\\hline
wordEnUni 	&	Unigram probability in English word-level language model.  The language model is built on English tweets.  If a word has a high probability of being English then it is likely English.  	&	-4	\\\hline
wordEnBi	&	Bigram probability in English word-level language model of the word with the word that precedes it.  If the probability is high then it is likely that it is an English word that follows another English word.	&	-4	\\\hline
charEnNgram	&	Trigram probability in English character-level language model of characters in a word.  This checks if it is likely sequence of characters in an English word.  	&	-2	\\\hline
charArNgram	&	Trigram probability in Arabizi character-level language model of characters in a word.  This checks if it is likely sequence of characters in an Arabizi word.	&	-13	\\\hline
 \end{tabular}
}
\end{center}
\caption{\label{features}Used labels for words}
\end{table*}

Table~\ref{id-res} reports on the language identification results and breaks down the results per word type and provides examples of mislabeling. Overall we achieved a word-level language identification accuracy of 98.5\%.  As the examples in the table show, the few mislabeling mistakes included: Arabized English words, Arabizi words that happen to be English words, single Arabizi words surrounded by English words (or vice versa), and misspelled English words.

\section{Arabizi to Arabic}
As mentioned earlier, Arabizi is simply Arabic, whether MSA or dialectal, that is written using Latin characters.  We were able to collecting Arabizi text by searching for common Arabizi words on Twitter, identifying the authors of these tweets, and then scraping their tweets to find more tweets written in Arabizi.  In all, we constructed a collection that contained 3,452 training pairs that have both Arabizi and Arabic equivalents.  All Arabizi words were manually transliterated into Arabic manually by a native Arabic speaker.  Some example pairs are:
\begin{itemize}
\item 3endek $\rightarrow$	<`ndk> (meaning ``in your care")
\item bytl3 $\rightarrow$ <by.tl`> (meaning ``he ascends")
\end{itemize}
For testing, we constructed a set of 127 random Arabizi tweets containing 1,385 word.  Again, we had a native Arabic speaker transliterate all tweets into Arabic script. An example sentences is:\\
\begin{itemize}
\item sa7el eih ? howa ntii mesh hatigi bokra $\rightarrow$ <sA.hl Ayh ? hw Anty m^s htyjy bkrT>
\item meaning:  what coast ? aren't you coming tomorrow
\end{itemize}

\begin{table*}[!ht]
\begin{center}
{\scriptsize
\begin{tabular}{|c|c|c|c|p{4.4cm}|p{5.3cm}|}
\hline
Actual Tag	&	Predicted Tag	&	Count	&	Percent	&	Example (Misclassified Token Highlighted)	&	Analysis	\\\hline
a	&	a	&	1909	&	99.1\%	&		&		\\\hline
a	&	e	&	12	&	0.6\%	&    tfker b2y {\bf shy} be relax, {\bf tab} 3 3ashan el talta tabta 	&    shy \& tab:  words that exist in English but are actually Arabic in context	\\
	&		&		&		&    al {\bf weekend} eljaay ya5i	 &  weekend:  Arabized English words	\\
	&		&		&		&    wow {\bf be7keelk} the cloud covered	&    bt7keelk: sudden context switch before and after	\\\hline
a	&	o	&	5	&	0.3\%	&	ya Yara ha call u @fjoooj eeeeeeeh	&	ha \& eeeeeeh: mistaken for smiles or laughs	\\\hline
e	&	e	&	773	&	97.0\%	&		&		\\\hline
e	&	a	&	21	&	2.6\%	&	el {\bf eye drope} eh ya fara7 	&	eye \& drop: sudden context switch	\\
	&		&		&		&	{\bf offtoschool}	&	offtoschool: misspelled English words	\\\hline
e	&	o	&	3	&	0.4\%	&	4 those going 2 tahrir	&	4 \& 2: numbers used instead of words	\\\hline
o	&	o	&	758	&	98.7\%	&		&		\\\hline
o	&	e	&	3	&	0.4\%	&	URL's and name mentions	&	Could be fixed with either a simple rule or more training data	\\\hline
o	&	a	&	7	&	0.9\%	&		&		\\\hline
 \end{tabular}
}
\end{center}
\caption{\label{id-res}Used labels for words}
\end{table*}

We applied the following preprocessing steps on the training and test data:
\begin{itemize}
\item We performed the following Arabic letter normalizations~\cite{Habash2010}:
\begin{itemize}
\item <Y> (alef maqsoura) $\rightarrow$ <y> (ya) 
\item <'A> (alef maad), <'a> (alef with hamza on top), and <'i> (alef with hamza on the bottom) $\rightarrow$ <A> (alef)
\item <|u'> (hamza on w), and <|Y'>  (hamza on ya) $\rightarrow$ <"'>  (hamza)
\item <T> (taa marbouta) $\rightarrow$ <h>  (haa)
\end{itemize}
\item Since people often repeat letters in tweets to indicate stress or to express emotions, we removed any repetition of a letter beyond 2 repetitions~\cite{Darwishetal2012}.  For example, we transformed the word ``salaaaam" to ``salaam".
\item Many people tend to segment Arabic words in Arabizi into separate constituent parts. For example, you may find ``w el kitab" (meaning ``and the book") as 3 separate tokens, while in Arabic they are concatenated into a single token, namely ``<wAlktAb>". Thus, we concatenated short tokens that represent coordinating conjunctions and prepositions to the tokens that follow them.  These tokens are: w, l, el, ll, la, we, f, fel, fil, fl, lel, al, wel, and b.
\item We directly transliterated the words ``isA" and ``jAk" to ``<'in ^sA' al-ll_ahi>" (meaning ``God welling") and to ``<jzAk al-ll_ahi xyrA>" (meaning ``may God reward you") respectively.
\end{itemize}

For training, we aligned the word-pairs at character level.  The pairs were aligned using GIZA++ and the phrase extractor and scorer from the Moses machine translation package~\cite{Koehnetal2007} . To apply a machine translation analogy, we treated words as sentences and the letters from which were constructed as tokens. The alignment produced letter sequence mappings. The alignment produced mappings between Latin letters sequences and Arabic letter sequences with associated mapping probabilities.  For example, here is a sample mapping:
\begin{itemize}
\item 2r $\rightarrow$ <qr> (p = 0.459) 
\end{itemize}

To generate Arabic words from Arabizi words, we made the fundamental simplifying assumption that any generated Arabic word should exist in a large word list.  Though the assumption fundamentally limits generation to previously seen words only, we built the word list from a large set of tweets.  Thus, the probability that a correctly generated word did not exist in the word list would be negligible.  This assumption allowed us to treat the problem as a mining problem instead of a generation problem where our task was to find a correct transliteration in a list of words instead of generating an arbitrary word.  We built the word list  from a tweet set containing a little over 112 million Arabic tweets that we scraped from Twitter between November 20, 2011 and January 9, 2012.  We collected the tweets by issuing the query ``lang:ar" against Twitter.  We utilized the tweet4j package for collection. The tweet set had 5.1 million unique words, and nearly half of them appeared only once.  

Our method involved doing two steps:\\

\textbf{Producing candidate transliterations:}  We implemented transliteration in a manner that is akin to the baseline system in El-Kahki et al.~\shortcite{ElKahkyetal2011}. Given an Arabizi word $w_{az}$, we produced all its possible segmentations along with their associated mappings into Arabic characters. Valid target sequences were retained and sorted by the product of the constituent mapping probabilities. The top $n$ (we picked $n$ = 10) candidates, $w_{ar_{1..n}}$ with the highest probability were generated.  Using Bayes rule, we computed:
	\begin{equation}
\underset{w_{ar_{i \in 1..n}}}{argmax}\:p(w_{ar_{i}}|w_{az}) = p(w_{az}|w_{ar_{i}}) p(w_{ar_{i}})
	\end{equation}
where $p(w_{az}|w_{ar_{i}})$ is the posterior probability of mapping, which is computed as the product of the mappings required to generate $w_{az}$ from $w_{ar_{i}}$, and $p(w_{ar_{i}})$ is the prior probability of the word.  \\


\textbf{Picking the best candidate in context:}  We utilized a large word language model to help pick the best transliteration candidate in context. We built a trigram language model using the IRSTLM language modeling toolkit~\cite{Federicoetal2008}.  The advantage of this language model was that it contained both MSA and dialectal text.  Given the top transliteration candidates and the language model we trained, we wanted to find the transliteration that would maximize the transliteration probability and language model probability.  Given a word $w_i$ with candidates $w_{i_{1-10}}$, we wanted to find $w_i \in w_{i_{1-10}}$ that maximizes the product of the transliteration probabilities (for all the candidates for all the words in the path) and the path probability, where the probability of the path is estimated using the trigram language model.
\begin{table}[h]
\begin{center}
{\scriptsize
\begin{tabular}{|c|c|c|c|p{3.7cm}|p{4.6cm}|}
\hline
rank	&	count	&	precentage	\\\hline
1	&	1,068	&	77.1\%	\\\hline
2	&	129	&	9.3\%	\\\hline
3	&	49	&	3.5\%	\\\hline
4	&	30	&	2.2\%	\\\hline
5	&	19	&	1.4\%	\\\hline
6	&	12	&	0.9\%	\\\hline
7	&	5	&	0.04\%	\\\hline
8	&	2	&	0.01\%	\\\hline
9	&	1	&	0.01\%	\\\hline
10	&	3	&	0.02\%	\\\hline
Not found	&	68	&	4.9\%	\\\hline
Total	&	1385	&		\\\hline
 \end{tabular}
}
\end{center}
\caption{\label{conv-res}Results of converting from Arabizi to Arabic with rank of correct candidates}
\end{table}

For testing, we used the aforementioned set of 127 random Arabizi tweets containing 1,385 word. We performed two evaluations as follows:\\

\textbf{Out of context evaluation.}  In this evaluation we wanted to evaluate the quality of the generated list of candidates.  Intuitively, a higher rank for the correct transliteration in the list of transliterations is desirable.  Thus, we used Mean Reciprocal Rank (MRR) to evaluate the generated candidates.  Reciprocal Rank (RR) is simply $\frac{1}{rank}$ of the correct candidate.  If the correct candidate is not in the generated list, we assumed that the rank was very large and we set RR = 0. MRR is the average across all test cases.  Notice that RR is 1 if the correct candidate is at position 1, 0.5 if correct is at position 2, etc.  Thus the penalty for not being at rank 1 is quite severe.\\
For out of context evaluation, we achieved an MRR of 0.84.  Table~\ref{conv-res} shows the breakdown of the ranks of the correct transliterations in the test set.  As can be seen, the correct candidate was at position one 77.1\% of the time.  No correct candidates were found 4.9\% of the time.  This meant that the best possible accuracy that we could achieve for in context evaluation was 95.1\%. Further, we examined the 68 words for which we did not generate a correct candidate.  Table~\ref{OOV} categorizes the 68 words (words are presented using Arabic script and Buckwalter encoding). Since Arabic dialects do not have a standard spelling convention, some of the word that we generated had a variant spelling from the ground truth.  Also in other cases, the correct morphological form did not exist in the word list or was infrequent.  In some of these cases, we generated morphologically related candidates that have an affix added or removed.  Some example affixes including coordinating conjunctions, prepositions, and feminine markers.

\begin{table}[h]
\begin{center}
{\scriptsize
\begin{tabular}{|p{2.8cm}|c|p{3.4cm}|}
\hline
Type	 &	Count & Examples \\\hline
no correct candidate	&	23	& \pbox{20cm}{wbenla2a7 ``and we hint to"\\- truth <wbnlq.h> wbnqH\\oleely ``tell me"\\- truth <qwlyly> qwlyly\\fsanya ``in a second"\\- truth <fy _tAnyT> fy vAnyp} \\\hline
spelling variant of word	&	17	& \pbox{20cm}{online ``online"\\- truth <AwnlAyn> AwnlAyn\\-guess <AnlAyn>AnlAyn\\betshoot ``you kick"\\- truth <bt^sw.t> bt\$wT\\-guess <bt^swt>bt\$wt} \\\hline
morphological variant	&	17	& \pbox{20cm}{bt7bii ``you (fm.) like"\\- truth <bt.hby> btHby\\-guess <bt.hbyn> btHbyn\\tesharadeeni ``you kick me out"\\- truth <t^srdyny> t\$rdyny\\-guess <t^srdyn> t\$rdyn} \\\hline
English word	&	4	& cute; mention; nation; TV\\\hline
no candidate generated	&	4	& \pbox{20cm}{belnesbalko ``for you"\\- truth <bAlnsblkm> bAlnsblkm\\filente5abat ``in the election"\\- truth <fAlAntxbAt> fAlAntxbAt} \\\hline
mixed Arabic \& English	&	3	& \pbox{20cm}{felguc ``in the GUC"\\- truth <fAl>GUC fAl-GUC\\ellive ``the live"\\- truth <Al>live Al-live}\\\hline
 \end{tabular}
}
\end{center}
\caption{\label{OOV}Analysis of words for which we did not generate candidates}
\end{table}

\textbf{In context evaluation.}  In this evaluation, we computed accuracy of producing the correct transliterated equivalent in context.  For in context evaluation, if we used a baseline that used the top out-of-context choice, we would achieve 77.1\% accuracy.  Adding a trigram language model, we achieved an accuracy of 88.7\% (157 wrong out of 1,385).  Of the wrong guesses, 91 were completely unrelated words and 46 were spelling or morphological variants.  

\section{Conclusion}
In this paper, we presented methods of detecting Arabizi that is mixed with English text and converting Arabizi to Arabic. For language detection we used a sequence labeler that used word and character level features.  Language detection was trained and tested on datasets that were constructed from tweets.  We achieved an overall accuracy of 98.5\%.  For converting from Arabizi to Arabic, we trained a transliteration miner that attempted to find the most likely Arabic word that could have generated an Arabizi word.  We used both character transliteration probabilities as well as language modeling.  We achieved 88.7\% transliteration accuracy. 

For future work, we would like to experiment with additional training data and improved language models that account for the morphological complexities of Arabic.  Also, the lack of spelling conventions for Arabic dialects may warrant detecting variant spellings of individual dialectal words and perhaps converting from dialectal text to MSA.

\end{document}